# From calibration to parameter learning: Harnessing the scaling effects of big data in geoscientific modeling

Wen-Ping Tsai[1], Dapeng Feng[1], Ming Pan[2,3], Hylke Beck[4], Kathryn Lawson[1,5], Yuan Yang[6,7], Jiangtao Liu[1], and Chaopeng Shen[1,5]*

[1] Civil and Environmental Engineering, Pennsylvania State University, University Park, PA, USA
[2] Center for Western Weather and Water Extremes, Scripps Institution of Oceanography, University of California San Diego, La Jolla, CA, USA
[3] Civil and Environmental Engineering, Princeton University, Princeton, NJ, USA
[4] GloH2O, Almere, the Netherlands
[5] HydroSapient, Inc., State College, PA, USA.
[6] Department of Hydraulic Engineering, Tsinghua University, Beijing, China
[7] Institute of Science and Technology, China Three Gorges Corporation, Beijing, China

* Corresponding author: Chaopeng Shen. **Email:**  cshen@engr.psu.edu

**Author Contributions**

WT ran the VIC-based experiments and wrote an early draft together with CS; DF ran the HBV-based experiments, MP ran some of the VIC model runs; JL provided assistance for data preparation; YY provided advice on running the VIC model; HB provided the global headwater catchment dataset and results from a regionalization scheme for comparison; CS conceived the study and revised the manuscript. MP, KL, DF, HB and CS edited the manuscript.

**This PDF file includes:**

Main Text
Figure 1 to 6, S1 to S4
Table S1

**Abstract**

The behaviors and skills of models in many geosciences, e.g., hydrology and ecosystem sciences, strongly depend on spatially varying parameters that need calibration. Here we propose a novel differentiable parameter learning (*dPL*) framework that solves a pattern recognition problem and learns a more robust, universal mapping. Crucially, *dPL* exhibits virtuous scaling curves not previously demonstrated to geoscientists: as training data collectively increases, *dPL* achieves better performance, more physical coherence, and better generalization, all with orders-of-magnitude lower computational cost. We demonstrate examples of calibrating models to soil moisture and streamflow, where *dPL* drastically outperformed state-of-the-art evolutionary and regionalization methods, or requires ~12.5% the training data to achieve the similar performance. The generic scheme promotes the integration of deep learning and process-based models, without mandating reimplementation.

**Main Text**
**Introduction**

This work broadly addresses geoscientific models across a wide variety of domains, including non-dynamical system models like radiative transfer modeling[1], as well as dynamical system models such as land surface hydrologic models (in earth system models) that simulate soil moisture, evapotranspiration, runoff, and groundwater recharge[2]; ecosystem models that simulate vegetation growth and carbon and nutrient cycling[3]; agricultural models that simulate crop growth[4]; and models of water quality[5] and human-flood interactions[6]. Besides scientific pursuits, these models fill the operational information needs for water supply management, pollution control, crop and forest management, climate change impact estimation, and many others.

A central and persistent problem concerning this wide variety of geoscientific models is that their behaviors and skills are strongly impacted by unobservable yet underdetermined parameters. But uncertainties in these parameters, such as for those controlling climate models estimating land surface feedbacks of water and $CO_2$ to the atmosphere, limit the confidence we have in the modeled results, such as simulated regional impacts caused by increasing $CO_2$ levels[7]. For decades, parameter calibration has been the orthodoxical procedure that is deeply entrenched across various geoscientific domains. A research enterprise and many textbook chapters have been dedicated to these calibration techniques and their applications in geosciences. Myriad methods including genetic algorithms[8–10] and evolutionary algorithms (EAs) such as the Shuffled Complex Evolutionary algorithm (SCE-UA)[11] have been introduced for calibration. For example, nearly all models for the rainfall-runoff process[12,13] (a key step in earth system models) and for ecosystem dynamics[14] involve unobservable parameters that require calibration. Moreover, these parameters are often sensitive to changes in spatial and/or temporal resolutions[15], other model parameters, model version, and input data, continuously triggering the need to readjust previously calibrated parameters - a repetitive and tedious process[16]. Current optimization algorithms require thousands of model runs or more, just to calibrate a dozen parameters.

Geoscientific processes have commonalities and dissimilarities between regions which could potentially be collectively leveraged by a big-data learning procedure. However, because traditional calibration procedures are generally applied to each location individually, they cannot exploit commonalities: in other words, they do not take advantage of what is learned in one place to apply it elsewhere. Because of the small amount of data at each site, algorithms may overfit to training data and find non-physical parameters, meaning they captured noise instead of a true signal. This often leads to dreaded non-uniqueness (i.e., equifinality)[17–19], where widely different parameter sets give similar evaluation metrics and thus cannot be determined by calibration. Site-by-site calibration often produces disparate, discontinuous parameters for neighboring, geographically-similar areas, as shown for land surface hydrologic models[16]. In summary, the traditional parameter calibration paradigm has become a bottleneck and a distinctive pain point to large-scale modeling in geoscientific research.

A class of techniques collectively referred to as parameter regionalization attempt to apply a more stringent constraint using all available data, which can help alleviate these issues[20]. A specific type of regionalization scheme prescribes transfer functions to relate known physical attributes to parameters[15,21]. The structures of the transfer functions are determined by humans (and thus need to be specifically customized for each model and data source), and the rigid form often limits efficacy in predicting parameters. As we will show, known regionalization schemes generate sub-optimal solutions that are fundamentally not ready to leverage big data. Also, they cannot handle a large number of parameters in the transfer functions, and are restricted to simple input attributes.

Recently, deep learning (DL)[22,23], a type of neural network with large depth, has shown great promise across scientific disciplines, including the geosciences[24–26], although some limitations also emerged. In the field of hydrology, previous work including ours has shown that a time series DL network model such as the long short-term memory (LSTM) algorithm[27] successfully learned satellite-sensed soil moisture dynamics, which control many processes such as infiltration, runoff, evapotranspiration, vegetation functioning, and drought resilience[28]. LSTM-based models have had success predicting soil moisture[29–31], streamflow[32], stream temperature[33], dissolved oxygen[34], and lake water temperature[35]. However, such a data-driven modeling method only allows for the prediction of observable variables for which we have sufficient data. For similarly important but unobserved prognostic variables such as evapotranspiration, groundwater recharge, or root carbon storage, we are still reliant on manually-calibrated process-based models (PBM).

Modern DL networks and their highly efficient training procedures, i.e., backpropagation and gradient descent, are well-suited to exploit the information in large datasets. One could envision leveraging DL to solve parameter calibration at large scales, but despite repeated calls to integrate physics and DL methods[24], there are no frameworks that exploit modern DL for the parameter calibration problem to the best of our knowledge. Most modern machine learning platforms support automatic differentiation (AD) which automatically keeps track of records all gradients, but traditional programming environments do not, and it will incur a huge expert labor cost to reimplement existing models on differentiable platforms. Without AD, the derivatives may also in theory be approximated by finite difference, but this is less accurate and computational intractable for a large neural network. While neural networks have been used in traditional parameter calibration, their typical, shallow role has been that of an efficient surrogate model, which emulates a PBM to provide reduced computational time during calibration[36]. With this paradigm, the calibration problems are still solved independently for each site. What *dPL* represents is a much deeper integration between DL and PBM.

Here we propose a differentiable parameter learning (*dPL*) framework based on deep neural networks, with two versions suitable for different use cases in geosciences. Our framework transforms the typically inverse parameter calibration problem into a big-data DL problem, leveraging the efficiency and performance of the modern DL computing infrastructure.

In the first and main case study, we applied the *dPL* framework to the Variable Infiltration Capacity (VIC) model, a widely-used land surface hydrologic model[37], producing parameters which allowed VIC to best match surface soil moisture observations from NASA's Soil Moisture Active Passive (SMAP) satellite mission[38]. We compared parameters from the *dPL* to those from SCE-UA, a standard evolutionary algorithm, and also to the operational parameters from the widely-used North American Land Data Assimilation System phase-II dataset (NLDAS-2)[39]. The comparisons were done at multiple training sampling densities, and different lengths of training data. We evaluated the quality of the estimated parameters for locations outside of the training set, and also for an uncalibrated variable, evapotranspiration (ET). In the second case study, we trained the framework on the CAMELS dataset[40], which consists of 531 basins in the US, with daily streamflow as the target and VIC as the PBM. We compared our results to the Multi-scale Parameter Regionalization (MPR) approach[21]. In the third case study, we trained on a global streamflow dataset for spatial extrapolation, or prediction in ungauged basins (PUB), in comparison to another state-of-the-art regionalization scheme recently published by Beck et al.[41]. We trained a separate neural network as a surrogate for VIC for the first two cases while we directly implemented simple conceptual hydrologic model, Hydrologiska Byråns Vattenbalansavdelning (HBV), in a DL platform for the third case.

## Results
**Optimization performance and efficiency**

For the SMAP calibration case study, our results (Figure 1) show that *dPL* can deliver equivalent or slightly better ending performance metrics than the evolutionary algorithm SCE-UA, at orders-of-magnitude lower cost, over the entire contiguous United States (CONUS). At training sampling densities of $1/8^2$ and $1/4^2$, where $1/8^2$ represents sampling one gridcell from each 8x8 patch (also abbreviated as s8, see Figure S1 in Supplementary Information), *dPL* had nearly identical ending error metrics (root-mean-square error, RMSE, between the simulated and observed surface soil moisture) as SCE-UA (Figure 1e-f). A slight advantage of *dPL* over SCE-UA manifested in correlation (Figure 2d), suggesting that the calibrated parameters captured seasonality trends well. *dPL*'s marginal outperformance (or virtual equivalence) in the median RMSE was a surprise to us, as one would expect an EA like SCE-UA to best capture the global minimum. This result attests to the uncompromising optimization capability of gradient descent. It also suggests there are commonalities to be leveraged in hydrologic processes across different sites. We observed that *dPL*'s performance is related to the amount of training data: it had the lowest performance (highest ending RMSE) when there was only 1 year of training data with the lowest sampling density (s16, or $1/16^2$).

Notably, *dPL* descended into the range of acceptable performance orders of magnitude faster than SCE-UA when given more training data (Figure 1e-f). For the model trained for 2 years, *dPL* required 310, 60, and 5 epochs to drop below the threshold for a functional model (RMSE=0.05) at $1/16^2$, $1/8^2$, and $1/4^2$ sampling densities, respectively, as compared to SCE-UA's 840 epochs (here an epoch for SCE-UA also means one forward simulation per gridcell; see Methods). With the identical surrogate model, at $1/8^2$ *dPL* required 9.9 minutes to achieve this RMSE level compared to SCE-UA's 253 minutes. An even more drastic difference of 11 minutes versus 1,004 minutes occurred at $1/4^2$ sampling density. The reduction of epochs resulted from the use of a domain-wide loss function and mini-batch training, allowing *dPL* to learn across locations rapidly (more interpretation in Discussion). In comparison, the sampling density had no effect on SCE-UA.

While using a surrogate is not novel, the efficiency of the LSTM surrogate model saved over two orders of magnitude of computational time as compared to the original VIC model. Training this surrogate (see Methods) required multiple iterations of CONUS-scale forward simulations. All things considered, the CONUS-scale calibration job which would have normally taken a 100-core CPU cluster to run 2-3 days with SCE-UA now only takes roughly 40 minutes on a single Graphical Processing Unit (GPU), representing a difference of roughly three to four orders of magnitude. While there are more efficient variants of SCE-UA, we compared dPL to the standard algorithm because it is well understood and benchmarked, and its variants do not differ by orders-of-magnitude in efficiency.

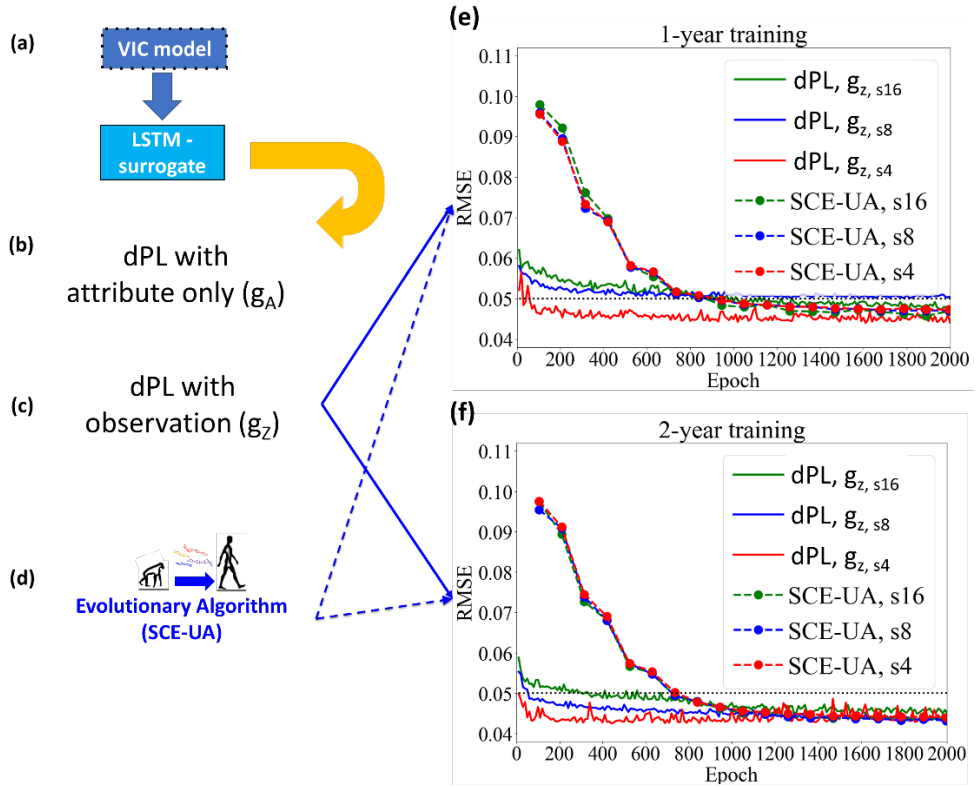

*Figure 1*. *Comparison of the dPL framework to the traditional calibration paradigm and the model performance as a function of training epochs. (a) An LSTM model is trained to mimic the outputs of a process-based land surface hydrologic model, VIC. (b) The forward-mapping parameter estimation framework (fPL) with network* gA*. (c) the* fPL *version with network gz. (d) Traditional site-by-site parameter calibration framework; (e) Objective function (RMSE) for the testing period vs. epoch (number of forward runs per pixel). Dashed lines are for SCE-UA and solid lines are for dPL. The models were trained with 1 year's worth of data. s16, s8, and s4 denote models trained with sampling densities of $1/16^2$, $1/8^2$, and $1/4^2$, respectively, where $1/16^2$ represents sampling one gridcell from each 16x16 patch. An epoch means, on average, one forward model run per gridcell for both dPL and SCE-UA. (f) Same as (e) but for models trained with 2 years' worth of data.*

**Spatial extrapolation and uncalibrated variables**

Here we show that *dPL* generalized better in space and gave parameter sets that were spatially coherent and better constrained, especially as the amount of data increased. Our spatial generalization test showed *dPL*'s metrics had almost no degradation from the training set to the out-of-training neighboring gridcells (Figure 2d). In contrast, SCE-UA's ubRMSE increased for the neighboring gridcells, with a statistically significant difference. More apparently, *dPL* had a much smaller spread of bias compared to SCE-UA (Figure 2d left panel). $g_z$ was slightly better than $g_A$, and both were better than SCE-UA in the spatial generalization test.

These comparisons suggest that *dPL* learned more robust parameter mapping pattern than SCE-UA, a strength we attribute to using the global constraint. The difference in metrics between SCE-UA and *dPL* was statistically significant and random seeds can be rejected as a cause (Table S1 in Supplementary Information). However, the difference may not appear large, which is to be expected as the difference was bounded by soil moisture physics, similar atmospheric inputs, and spatial proximity to training sites. A larger difference is noted in Example 3 below.

Most geoscientific models output multiple unobserved variables of interest that can be used as diagnostics or to support narratives of the simulations. It can be argued that if a parameter set leads to improved behavior for both calibrated (soil moisture) and uncalibrated (evapotranspiration, ET) variables, it better describes the underlying physical processes, and the model *gives good results for the right reason*. Here we compared model-simulated temporal-mean ET to MOD16A2, an ET product from the completely independent Moderate Resolution Imaging Spectroradiometer (MODIS) satellite mission (see Methods for discussion of limitations).

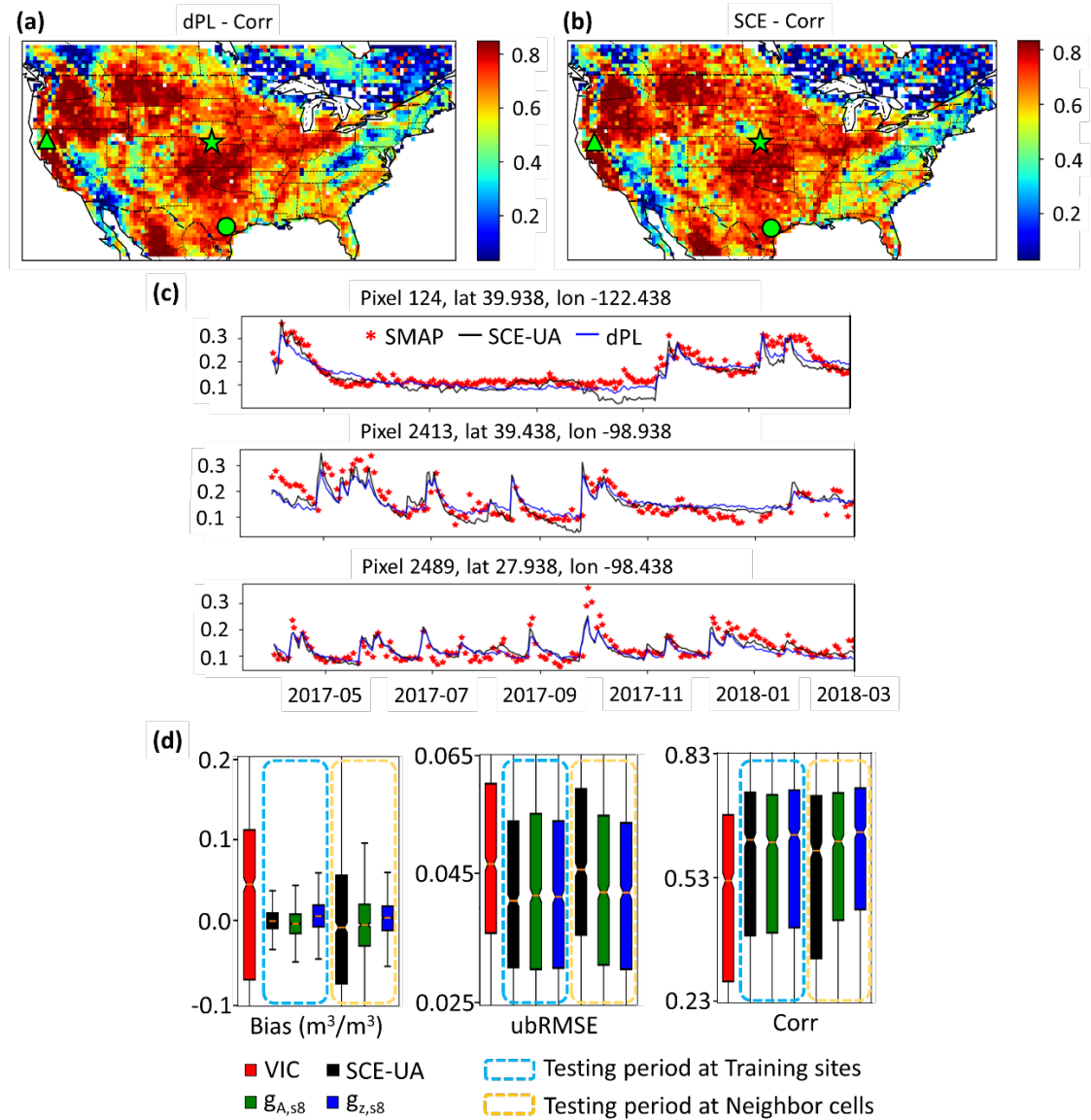

*Figure 2. Performance of SMAP simulations. (a) Maps of correlations between observed and simulated soil moisture during test period at $1/4^2$ sampling density. (b) Same as (a) but from SCE-UA. (c) Example time series at randomly selected sites. (d) Boxplots to summarize metrics from gridcells for temporal generalization test (evaluated on the training locations over the test period,*

*blue dashed box) and spatial generalization test (orange dashed box) at $1/8^2$ sampling density (one gridcell from an 8x8 patch, see illustration in Figure S1 in the Supplementary Information) for VIC (NLDAS-2 default parameters), SCE-UA, $g_z$, $g_A$. In the spatial generalization test, we also sampled at $1/8^2$ sampling density for training and evaluated the parameters' performance on a neighbor 3 rows to the north and 3 columns to the east from each of the training gridcells, over the test period. We tested on other neighboring gridcells as well, with similar results.*

The parameters from *dPL* produced ET that was closer to MOD16A2 in spatial pattern than did either the parameters from NLDAS-2 or those calibrated by SCE-UA (Figure 3). At $1/8^2$ sampling density, the CONUS-median values for correlation between observed and simulated ET for $g_z$, and SCE-UA was 0.75 (ensemble mean) and 0.69 (ensemble mean), respectively, and the differences were multiple standard deviations (due to random seeds) apart (Table S1 in supplementary information). Due to the much smaller bias, the Nash-Sutcliffe model efficiency coefficient (NSE) for *dPL* ($g_z$) was 0.55, as opposed to 0.38 for SCE-UA and 0.44 for NLDAS-2. SCE-UA calibration using soil moisture did not improve the spatial pattern of ET compared to the current parameter sets in NLDAS-2, but *dPL* led both by a substantial margin. Similar to soil moisture, ET variation is strongly driven by rainfall and energy inputs, so we should not expect the model to give wildly worse results even if parameter sets are not ideal.

While MOD16A2 should not be considered truth, it utilizes completely separate sources of observations including leaf area index and photosynthetically active radiation (see Methods). Thus, the better agreement of *dPL* with MOD16A2, in terms of both correlation and bias, suggest that *dPL* had more physically-relevant parameter sets. When SCE-UA calibrated a certain gridcell, it did not put this gridcell in the context of regional patterns, so it could sometimes distort model physics in its pursuit of lowest RMSE in soil moisture for that location. For *dPL*, because the inputs to the parameter estimation module, i.e., forcings, responses, and attributes, are themselves spatially coherent (autocorrelated), and only one uniform model is trained, the inferred parameters are also spatially coherent.

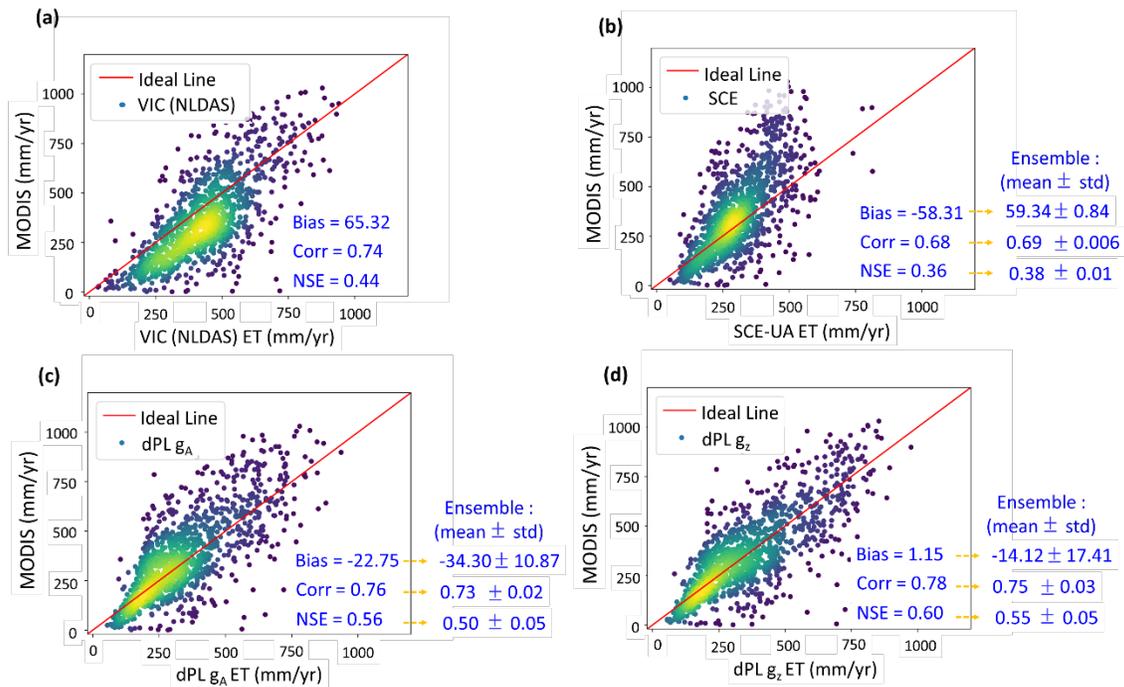

***Figure 3.*** *Uncalibrated variable (ET) metrics from models trained at $1/8^2$ sampling density. Scatter plots of temporal-mean ET (mm/year) comparing (a) MOD16A2 with ET produced by (b) NLDAS-2, (c) SCE-UA, and (d) dPL. Each point on the scatter plot is the temporal mean ET of a ⅛-latitude-longitude-degree gridcell defined on the NLDAS-2 model grid. Yellow color indicates higher density of points. The ensemble metrics are from training the model with different random seeds, while the 1-vs-1 plots came from one particular random realization. Panel (a) is for the NLDAS-2's default VIC simulation and does not have an ensemble.*

The inferred parameter fields reveal the reason behind the advantage of *dPL* over SCE-UA. One of the parameters estimated by *dPL, INFILT* (see Methods), shows a spatial pattern that generally follows the aridity and topographic patterns of the CONUS (Figure 4a), which agrees with our general understanding of physical hydrology and the VIC model's behavior. Precipitation declines from east to west until reaching the Rocky Mountains, and then increases again at the west coast. The large southeast and northwest coasts are the wettest parts of the country due to moisture from the oceans. *dPL* kept the steepness of the infiltration capacity curve of surface runoff smooth in wet areas to produce more runoff, which is consistent with earlier literature[16,42]. INFILT varies continuously in the southeast coastal plains where soil is thick and permeable and most rainfall infiltrates[43]. The map also captured patterns of poorly drained soils to the northeastern edge of the map, which are visible from soil surveys. We observed such continuity with other parameters as well (Figure S3 in Supplementary Information).

In contrast, SCE-UA (Figure 4b) presented discontinuous parameters apparently plagued by stochasticity and parameter non-uniqueness, which explains why SCE-UA had worse performance in the spatial generalization and uncalibrated variable tests. Soil moisture observations impose constraints only on a part of the system, and VIC, like any PBM, also contain structural deficiencies; therefore, it is unreasonable to expect *dPL* (or SCE-UA, or any other scheme) to fully remove parametric uncertainties or find the most realistic parameters. Nevertheless, the parameters found by *dPL* seem more coherent with known physical relationships than those from SCE-UA.

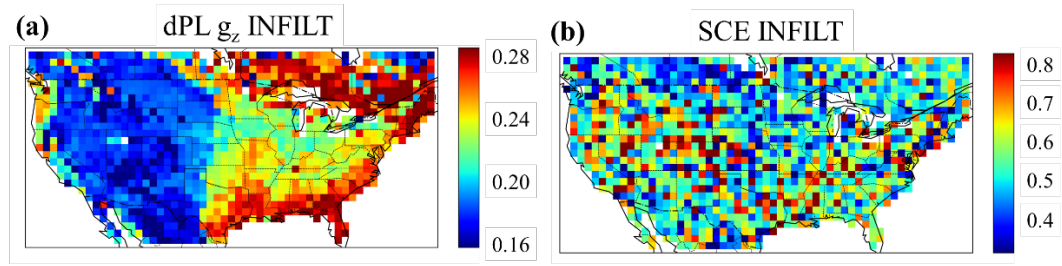

*Figure 4. Comparison of parameters generated by dPL and SCE-UA. The continuous, spatially-representative patterns of (a) dPL-inferred parameters are noteworthy, especially in comparison to the discontinuous, random appearance of (b) SCE-inferred parameters from site-by-site calibration. Both were trained with a $1/8^2$ sampling density.*

**Streamflow cases in comparison with regionalization schemes**
The streamflow temporal generalization showed even more pronounced advantages of dPL over existing state-of-the-art multiscale parameter regionalization (MPR). In the case with the CAMELS hydrologic dataset (Figure S4a), we applied *dPL* to estimate parameters for the VIC hydrologic model and tested the parameters in a different period of time (temporal generalization) on basins in the training set. $g_z$ achieved a median NSE of ~0.44, compared to the median value of 0.32 reported for MPR[21] (Figure 5a). This result challenges the previous argument that a low value of 0.32 in this experimental setup was close to the performance ceiling of VIC due to its structural deficiencies and showed that the regionalization scheme was also not near optimal. Meanwhile, a pure LSTM model can certainly achieve higher NSE than VIC[32], suggesting there is substantial room to improve VIC structure.

A large advantage of *dPL* was also noticed in the global PUB case (spatial extrapolation), we applied *dPL* with two sets of input features on a global hydrologic dataset (Figure S4b) and tested on basins not included in the training set. An existing state-of-the-art regionalization scheme from Beck et al.[41], (hereafter referred to as the Beck20 scheme) reported a median Kling-Gupta efficiency coefficient (KGE, similar in magnitude to NSE, see Methods) of 0.48[41], while dPL gave values of 0.56 for the comparable 8-feature setup (Figure 5b). We witnessed a noticeable separation of the cumulative distribution function (CDF) curves between *dPL* and Beck20 throughout the different ranges of KGE. In addition, the 27-feature *dPL* setup generalized better in space than the 8-feature *dPL* setup (median KGE=0.59), suggesting using more attributes as inputs did not cause *dPL* to overfit. It would take considerably more effort for the traditional schemes to run the 27-feature setup as it entails including more transfer functions and more parameters to train.

These differences of around 0.1 in median NSE or KGE in both of temporal and spatial experiments are quite significant, as NSE=1 indicates a perfect model while NSE=0 corresponds to using the mean value as the prediction. Increases in NSE from 0.32 to 0.55 in the VIC case, or from 0.48 to 0.56 in the HBV case represent material changes in model reliability for water management planning applications. It is worth noting that our results are still limited by the structure of PBMs and the ways the reference problems were setup in the literature, e.g., choices of inputs (not using climate attributes), no routing in the global PUB case, etc., as the main purpose here is to compare with previous schemes. While traditional regionalization schemes are an important and constructive avenue, these comparisons suggest that they are far from optimal and thus cannot fully leverage information provided by big data.

*dPL* offers unprecedented flexibility to leverage all forms of available information. It is not possible for regionalization schemes like MPR to map from time series. For soil moisture, our tests showed that using $g_z$ improved the simulations with statistical significance compared to using the attributes alone with $g_A$ for soil moisture, ET, and neighboring gridcell soil moisture (Table S1 in Supplementary Information).

**Data scaling curves**

Summarizing the results in another way, as amount of training data increases, we clearly witness virtuous scaling curves (Figure 6a), which, to our knowledge, have never been discussed in the context of geoscientific modeling. As training data increases, the performance (based on the ending RMSE for soil moisture), physical coherence (based on the uncalibrated variable ET), and generalization (based on spatial extrapolation) all improve, while the average cost per site decreases dramatically (based on training epochs). The reduction in cost can be interpreted as *dPL* demonstrating economies of scale (EoS), where a mildly rising global training cost is shared by all sites. But beyond EoS, the improvements in parameter performance and physical coherence indicate that each site now benefits from a better "service" resulting from the participation of other sites. Each additional data point allows the data-driven scheme to better capture details of the underlying parameter-response function, and more data imposes a stronger large-scale physiographical constraint that must be simultaneously satisfied, suppressing overfitting and improving robustness. This scaling effect is an important reason why *dPL* can surpass SCE-UA because at the low data-density end (s16 in Figure 1), *dPL*'s ending RMSE in fact was not as strong as SCE-UA.

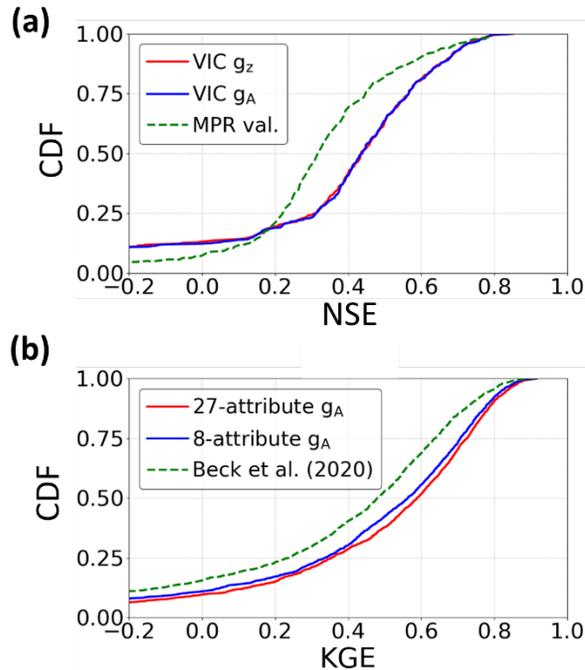

***Figure 5.*** *Comparison of dPL and regionalization schemes for streamflow calibration. (a) Calibrating the VIC hydrologic model via a surrogate model on the CAMELS dataset on CONUS, in comparison to MPR; (b) calibrating the HBV hydrologic model (not a surrogate) on the Beck20 global dataset, in comparison to the Beck20 regionalization scheme.*

For the global PUB experiments, we witness a two-phase scaling curve and strong data efficiency with *dPL* when we systematically reduced training data (Figure 6b). In the first phase (2%-25% of basins used in training), there is a rapid rise where the median KGE improved from 0.38 to 0.54. In the second phase, the improvement became much slower but did not saturate at 100% of the dataset. There is an upper bound to the PBM performance due to model structural deficiencies, so the slowdown is to be expected, but the initial scaling curve is surprisingly steep. We suspect the first rapid-rise phase is due to *dPL*'s ability to learn across sites (at 25% basin density, the model mostly learned the main characteristics of the problem), while the second, more asymptotic phase is due to reduced geographic distance between training and test basins (the model does fine tuning).

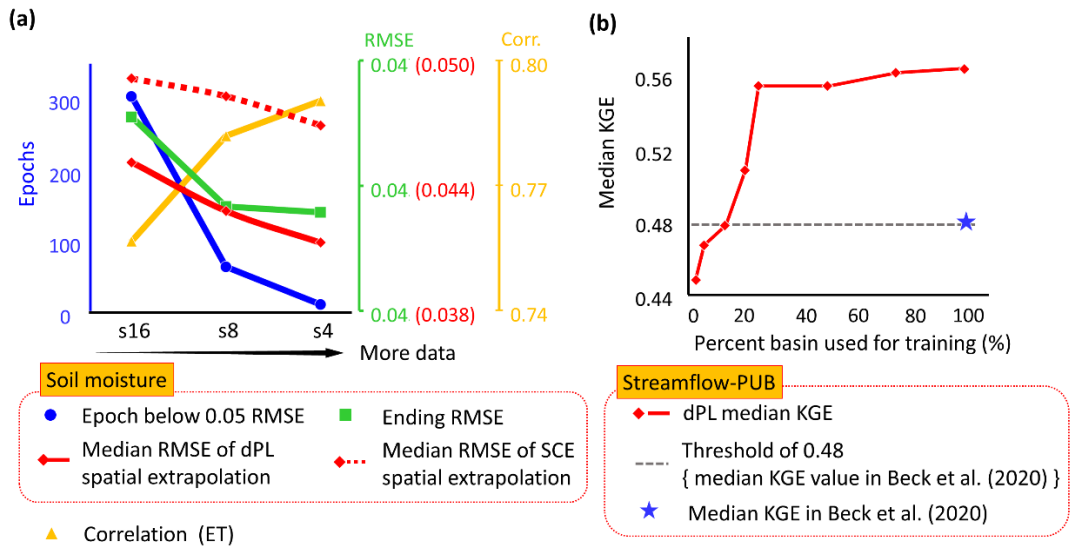

*Figure 6. Scaling curves of the dPL with respect to training data. (a) Scaling curve for soil moisture calibration as the sampling density for training increases. Lines are matched with the axes with the same color. Blue: the epoch for the calibration to drop below RMSE=0.05; Yellow: the spatial correlation between simulated ET (not calibrated against) and MODIS ET product; Green: the ending RMSE of soil moisture of learning; Solid Red: the RMSE of soil moisture or neighboring gridcells not included in training from dPL; Dashed red: the RMSE of soil moisture for neighboring gridcells not included in training from SCE-UA. (b) The scaling curve for the spatial extrapolation (PUB) test with the Beck20 global headwater catchment dataset.*

Overall, if we assume the Beck20 scheme to also improve monotonously with more basins on the PUB experiment (this is likely as more training basins mean smaller distance from test basins, akin to the second phase describe above), *dPL* would achieve the same performance as Beck20 (median ~KGE=0.48) using just ~12.5% of the training data.

## Discussion

There are several major implications of the results, as discussed below. First, the novel scaling curves, which we expect to hold true for other geoscientific domains, showed that *dPL*'s advantages largely arise from leveraging big data and the process commonalities and differences found therein.

The immense, simultaneous benefits of data scaling with respect to performance, efficiency and (unique to parameter learning) physical coherence had not yet been demonstrated in geosciences and some other domains. Both schemes of *dPL* demonstrated orders-of-magnitude savings in computational time over SCE-UA even with a moderate sampling density ($1/8^2$), enabling new scientific pursuits. Our results give geoscientists strong motivation to rise above case-specific datasets (which is currently a common practice even among machine learning studies, based on our review[25]), compile large datasets, and collectively ride such virtuous scaling curves up. The curves also suggest that any interpretation of DL model results must be grounded in the context of training data amount, e.g., comparisons involving DL in a small-data setting may have limited significance.

Second, *dPL* demonstrates the framework is especially helpful where a full simulation history involving unobserved variables is needed to provide interpretable narratives or diagnostics, e.g., as needed in climate change assessment applications. On the other hand, our approach can immediately and greatly boost the accuracy of large-scale, socio-economically important geoscientific models such as the National Water Model[44], which is responsible for predicting floods on a national scale. Following this integration, there should be many pathways towards leveraging machine learning to improve our physical understanding such as learning about better structures in the model.

Third, DL-supported *dPL* offers a generic, automatically adaptive and highly efficient solution to a large variety of models in geosciences and beyond. For our three examples, each with some different configurations, we used the same $g_A$ and $g_Z$ network components with little customization. We expect such genericity to carry over to other domains. This is because we do not explicitly specify any transfer functions: these are determined by the DL algorithm, which can adapt to new problems automatically. In contrast, for regionalization, new transfer functions need to be conceived for every new model, new calibrated parameters, or even new experiments, as shown in this work with 8-attribute vs 27-attribute model versions. *dPL* is more flexible than traditional methods to stand up to the challenges of widely different datasets and problem formulations.

Related to the above point, $g_A$ and $g_z$ each have their use cases. Ideally, if the PBM has high fidelity to reality and high-quality data $A$ can capture the main heterogeneities in processes, $g_A$ should be sufficient. In contrast, $g_z$ should be useful for other problems where models have deficiencies or there are latent variables not captured by attribute. An example is ecosystem modeling, where we have ample observations of top-canopy variables such as leaf area index, but forest species, successional stages, and understory communities lack detailed data except at a small number of sites. In these cases, we suspect that using $z$ in our $g_z$ will provide more pronounced benefits in accuracy and computational demand.

Depending on the setup, imposing physical constraints has been shown to improve generalization[35,45] and certainly builds an important bridge between process knowledge and data science. Overall, the DL-based *dPL* approach showed immense advantages in efficiency, performance, and robustness over traditional methods.

## Methods

### The process-based hydrologic model and its surrogate

The Variable Infiltration Capacity (VIC) hydrologic model has been widely used for simulating the water and energy exchanges between the land surface and atmosphere, along with related applications in climate, water resources (e.g., flood, drought, hydropower), agriculture, and others. The model simulates evapotranspiration, runoff, soil moisture, and baseflow based on conceptualized bucket formulations. Inputs to the model include daily meteorological forcings, non-meteorological data, and the parameters to be determined. Meteorological forcing data include time series of precipitation, air temperature, wind speed, atmospheric pressure, vapor pressure, and longwave and shortwave radiation. More details about VIC can be found in Liang et al.[37].

An LSTM was trained to reproduce the behavior of VIC as closely as possible. In theory, if a hydrologic model can be written into machine learning platforms (as in our HBV case), this step is not needed, but training a surrogate model is more convenient when the model is complex. If we did not use the surrogate model, SCE-UA would also have needed to employ the $O(10^2)$ more expensive CPU-based VIC model. We evaluated the accuracy of the surrogate model, and the median correlations between VIC and the surrogate simulation were 0.91 and 0.92 for soil moisture and ET, respectively (Figure S2 in Supplementary Information). This framework was implemented in the PyTorch deep learning framework[46].

**The long short-term memory (LSTM) network**

The long short-term memory network (LSTM) was originally developed in the artificial intelligence field for learning sequential data, but has recently become a popular choice for hydrologic time series data[26]. As compared to a vanilla recurrent neural network with only one state, LSTM has two states (cell state, hidden state) and three gates (input gate, forget gate, and output gate). The cell state enables long-term memory, and the gates are trained to determine which information to carry across time steps and which information to forget. These units were collectively designed to address the notorious DL issue of the vanishing gradient, where the accumulated gradients would decrease exponentially along time steps and eventually be too small to allow effective learning[47]. Given inputs *I*, our LSTM can be written as the following:

Input transformation: $\quad x^t = ReLU(W_I I^t + b_I) \quad$ (1)

Input node: $\quad g^t = tanh(\mathcal{D}(W_{gx} x^t) + \mathcal{D}(W_{gh} h^{t-1}) + b_g) \quad$ (2)

Input gate: $\quad i^t = \sigma(\mathcal{D}(W_{ix} x^t) + \mathcal{D}(W_{ih} h^{t-1}) + b_i) \quad$ (3)

Forget gate: $\quad f^t = \sigma(\mathcal{D}(W_{fx} x^t) + \mathcal{D}(W_{fh} h^{t-1}) + b_f) \quad$ (4)

Output gate: $\quad o^t = \sigma(\mathcal{D}(W_{ox} x^t) + \mathcal{D}(W_{oh} h^{t-1}) + b_o) \quad$ (5)

Cell state: $\quad s^t = g^t \odot i^t + s^{t-1} \odot f^t \quad$ (6)

Hidden state: $\quad h^t = tanh(s^t) \odot o^t \quad$ (7)

Output: $\quad y^t = W_{hy} h^t + b_y \quad$ (8)

Additional explanation of LSTM equations and variables can be found in our previous paper[32]. The LSTM network and our whole workflow were implemented in PyTorch[46], an open source machine learning framework.

**The network**

The whole network is trained using gradient descent, which is a first-order optimization scheme while some second-order schemes like Levenberg-Marquardt often have large computational demand and are thus rarely used in modern DL[48]. As with most DL work, the hyperparameters of *dPL* needed to be adjusted. We manually tuned hidden sizes, batch size, test-period RMSE, and train-test difference in RMSE as guidelines for selecting these hyperparameters. Higher sampling densities led to larger training data, which could be better handled by larger hidden sizes.

Based on Troy et al.[42], the calibrated parameters include the infiltration curve (INFILT), maximum base flow velocity (Dsmax), fraction of maximum base flow velocity where nonlinear base flow begins (Ds), fraction of maximum soil moisture content above which nonlinear baseflow occurs (Ws), and variation of saturated hydraulic conductivity with soil moisture (EXPT). INFILT decides the shape of the Variable Infiltration Capacity (VIC) curve and denotes the amount of available infiltration capacity. The formula regarding INFILT in VIC is shown as below:

$$i = i_m \left[ 1 - (1 - a_f)^{\frac{1}{INFILT}} \right] \quad (9)$$

where $i$ is infiltration capacity, $i_m$ is maximum infiltration capacity (related to the thickness of the upper soil layer in th), and $a_f$ is the fraction of saturated area. With other parameters and $a_f$ being the same, larger INFILT leads to a reduced infiltration rate and thus higher runoff.

**Satellite-based estimates of ET**

MOD16A2[49] is an 8-day composite ET product at 500-meter resolution, which is based on the Penman-Monteith equation. With this algorithm, MODIS 8-day fraction of photosynthetically-active radiation is used as the fraction of vegetation cover to allocate surface net radiation between soil and vegetation; MODIS 8-day albedo and daily meteorological reanalysis data are used to calculate surface net radiation and soil heat flux; and MODIS 8-day leaf area index (LAI) and daily meteorological reanalysis data are used to estimate surface stomatal conductance, aerodynamic resistance, wet canopy, soil heat flux, and other environmental variables. MODIS land cover is used to specify the biome type for each pixel to retrieve biome-dependent constant parameters.

We did not use MOD16A2 as a learning target; the purpose here was to validate which calibration strategy leads to better descriptions of overall model dynamics. MOD16A2 is not perfect, but since these are completely independent observations from those by SMAP, better agreement should still indicate better modeling of physics overall.

**Shuffled Complex Evolution (SCE-UA) for comparison**

For comparing the parameter estimation module in *dPL*, the Shuffled Complex Evolution (SCE-UA) method[11] introduced three decades ago but still relevant today[50], was also implemented as a reference method. We chose SCE-UA for comparison because it is well-established and widely applied. The algorithm ranks a population based on the objective function, and partitions a population of parameter sets into multiple subpopulations called complexes. In one iteration of SCE-UA, the complexes are evolved individually for a number of competitive evolution steps, where reflection, contraction, and random trials are attempted, before they are shuffled and redivided into new complexes for the next iteration.

To enable comparison, similar to *dPL*, we defined an epoch for SCE-UA as on-average one forward simulation for each gridcell. Hence one SCE-UA iteration contains many epochs. Because SCE-UA uses discrete iterations involving an uneven number of simulations/epochs, it was not as meaningful to compute the best objective function at the end of fixed epochs. Instead, across different gridcells, we collected the lowest objective function RMSE achieved from the beginning to the end of each iteration, and took the average of the ending epochs as the epoch to report. We tuned the number of complexes of SCE-UA and set it to seven.

**CAMELS streamflow test**

In earlier work, Mizukami et al.[21] calibrated the VIC model using the multiscale parameter regionalization scheme using data from 531 basins in the Catchment Attributes and Meteorology for Large-Sample Studies (CAMELS) dataset over CONUS (see basin locations in Figure S4a in Supplementary Information). They limited their scope to basins smaller than 2,000 km$^2$ and trained and tested on the same basins, making the experiment a test on the model's ability for temporal generalization. The calibration period of MPR was from 1 October 1999 to 30 September 2008, and the validation period was from 1 October 1989 to 30 September 1999. They calibrated transfer functions for 8 default VIC parameters and added two additional parameters (shape and timescale) for routing. To compare with Mizukami et al.[21], we used the same dataset, same basins and same training and test periods. They reported a median NSE of 0.30 for VIC.

**Global PUB test**

Beck20 presented a global-scale hydrologic dataset containing forcings, static attributes and daily streamflow data from 4,229 headwater basins across the world. They used a state-of-the-art regionalization scheme for prediction in ungauged basins (PUB), in which no data from test basins were used in the training dataset, thus testing the scheme's capability to generalize spatially. Eight attributes were used for the transfer functions in Beck20, including humidity, mean annual precipitation, mean annual potential evaporation, mean normalized difference vegetation index (NDVI), fraction of open water, slope, sand, and clay. They trained linear parameter transfer functions from raw predictors to 12 free parameters of a simple hydrologic model, HBV. 4,229 basins were divided into three climate groups: (i) tropical, (ii) arid and temperature, and (iii) cold and polar and trained transfer functions for each of these groups. They ran a cross validation within each group, e.g., for the arid and temperature group, they further divided the data into 10 random folds, trained the transfer functions in 9 of the 10 folds, and tested the transfer functions on the 10-th holdout fold; then they rotated to other folds as holdout data and reported the average metrics from these holdout basins. Because the holdout basins were randomly selected, there will always be neighboring basins that were included in the training set. However, the sparser the training data are, the further away the holdout basins will be, on average, from the training basins. Hence, we can reduce the number of training basins to examine the impacts of less training data on the model's ability to generalize in space.

Since the primary purpose of this work is to compare *dPL* to Beck20's regionalization scheme, we kept the setup as similar as possible and focused on the comparison with their temperate catchment group (see their locations in Figure S4b in Supplementary Information), for which they reported a median Kling Gupta model efficiency (KGE, see below) coefficient of 0.48. We similarly ran the 10-fold cross validation by training $g_A$ on 9 folds and testing it on the 10-th, and rotated the holdout fold. The training and test periods were both 2000-2016, the same as Beck20. We compared using 8 raw attributes (to be comparable to Beck20) as well as using the whole 27 attributes to demonstrate the extensibility of the *dPL* scheme.

**Evaluation metrics**

Four statistical metrics are commonly used to measure the performance of model simulation: bias, correlation (Corr), unbiased RMSE (ubRMSE), and Nash Sutcliffe model efficiency coefficient (NSE). Bias is the mean difference between modeled and observed results. ubRMSE is the RMSE calculated after the bias (systematic model error) is removed during the calculation, and measures the random component of the error. Corr assesses if a model captures the seasonality of the observation, but did not care about the correlation. NSE also considers bias and it is 1 for a perfect model and can be negative for poor models. $\bar{y}$ is the average modeled value of all pixels, and $\overline{y*}$ is the average observed value of all pixels.

$$Bias = \frac{\sum_{i=1}^{n}(y_i - y_i^*)}{n} \qquad (10)$$

$$ubRMSE = \sqrt{\frac{\sum_{i=1}^{n}[(y_i - \bar{y}) - (y_i^* - \overline{y*})]^2}{n}} \qquad (11)$$

$$Corr = \frac{\sum_{i=1}^{n}[(y_i - \bar{y})(y_i^* - \overline{y*})]}{\sqrt{\sum_{i=1}^{n}[(y_i - \bar{y})^2]}\sqrt{\sum_{i=1}^{n}[(y_i^* - \overline{y*})^2]}} \qquad (12)$$

$$NSE = 1 - \frac{\sum_{i=1}^{n}(y_i^* - y_i)^2}{\sum_{i=1}^{n}(y_i^* - \overline{y*})^2} \qquad (13)$$

All metrics were reported for the test period. When we evaluated $g_z$ on the training locations, historical observations during the training period ($z_{t,t \in T}$) were included in the inputs. Then, we used the parameters calibrated from the training period to run the model in the test period to get the reported metrics.

For the spatial generalization tests, we sampled one gridcell out of each 8x8 patch (sampling density of 1/8² or s8), and we ran the trained *dPL* model on another gridcell in the patch (3 rows to the north and 3 columns to the east of the training gridcell). For SCE-UA, we took the parameter sets from the nearest trained neighbor. For $g_A$, we sent in static attributes from the test neighbor to infer to the parameters, which is evaluated over the test period. For $g_z$, we sent in static attributes as well as forcings and observed soil moisture from the test neighbor, but from the training period. All models are evaluated over the test period.

To be comparable to Beck20, we also calculated the Kling-Gupta model efficiency coefficient (KGE), which is similar in magnitude to NSE:

$$KGE = 1 - \sqrt{(r-1)^2 + (\beta-1)^2 + (\gamma-1)^2} \quad (14)$$

$$\beta = \frac{\mu_s}{\mu_o} \text{ and } \gamma = \frac{\sigma_s/\mu_s}{\sigma_o/\mu_o} \quad (15)$$

where $r$ is the correlation coefficient between simulations and observations, $\beta$ and $\gamma$ are the bias and variability ratio, respectively. $\mu$ and $\sigma$ are the mean and standard deviation of runoff, and the indices s and o represent simulations and observations.

**Supplementary Information**

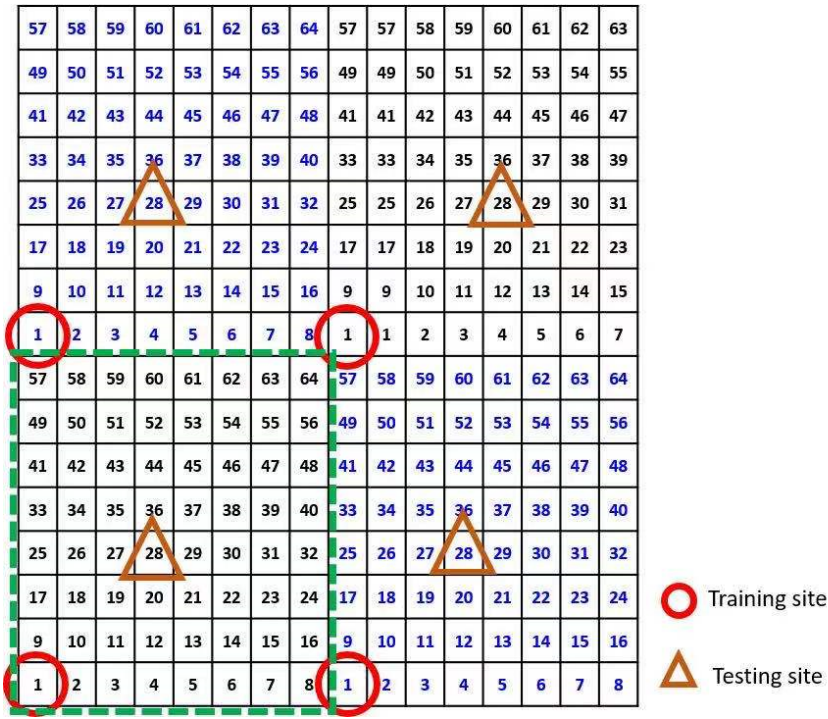

*Figure S1. An illustration of the sampling strategy and the training and test sites at density s8.*

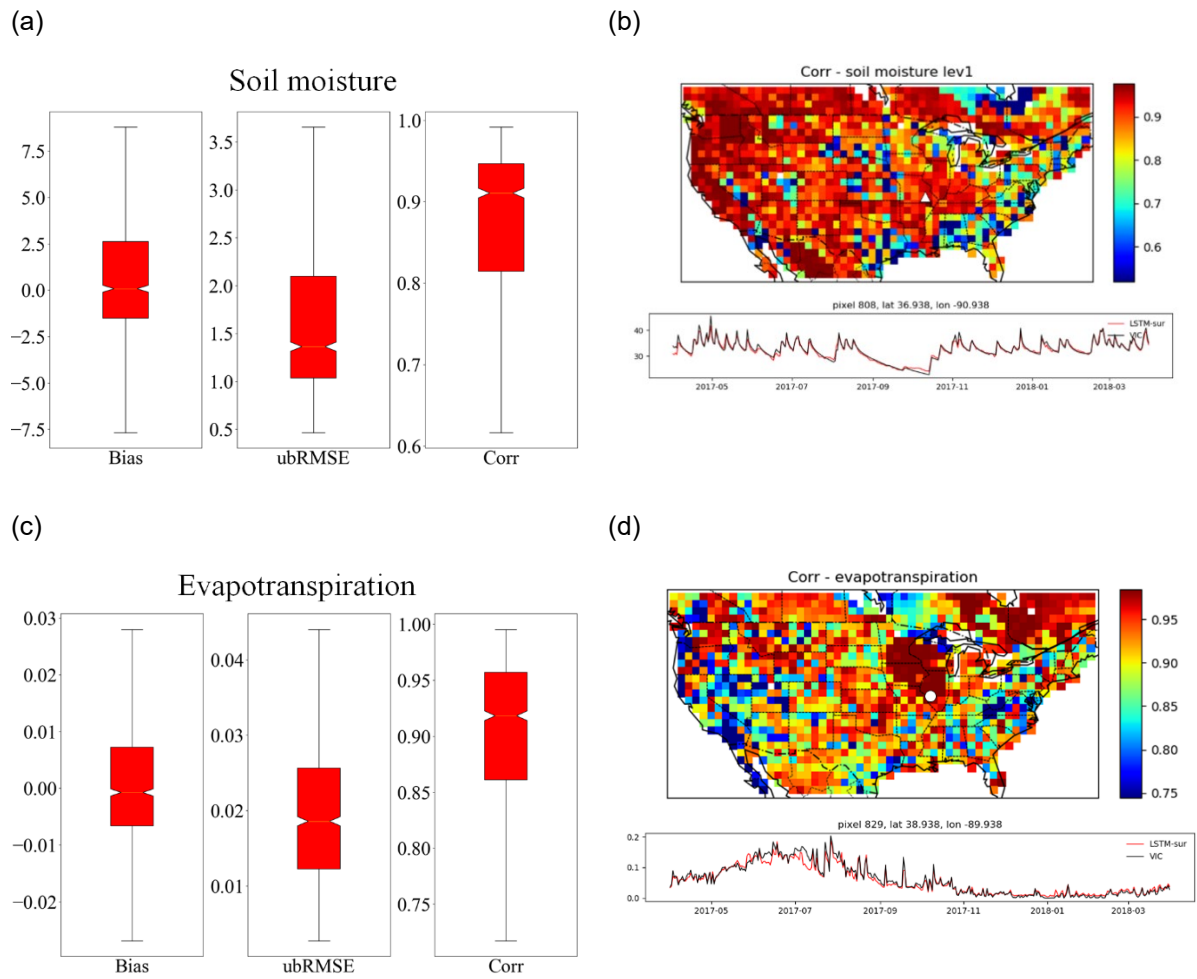

*Figure S2.* *The performance of the LSTM surrogate model (LSTM-sur). (a) Test metrics for the soil moisture of LSTM-sur. (b) Map of correlation between soil moisture from LSTM-sur and VI, and time series comparison at a gridcell. Black line denotes simulation of LSTM-sur. Red line denotes VIC. (c) Test metrics for the evapotranspiration of LSTM-sur as compared to VIC outputs. (d) Map of correlation between evapotranspiration from LSTM-sur and VIC; and time series comparions at a gridcell for evapotranspiration.*

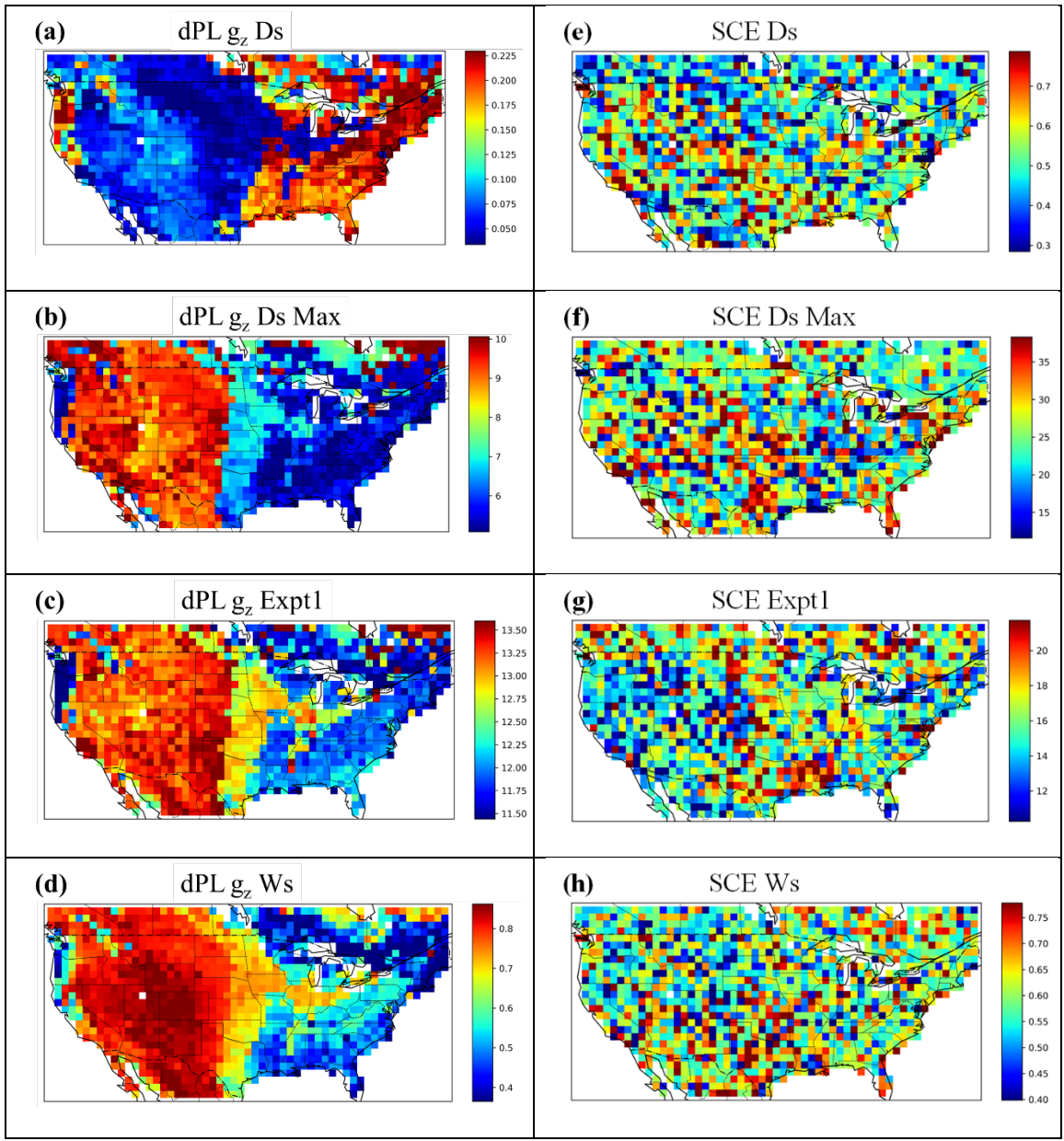

***Figure S3.*** *Maps of parameters other than INFILT.*

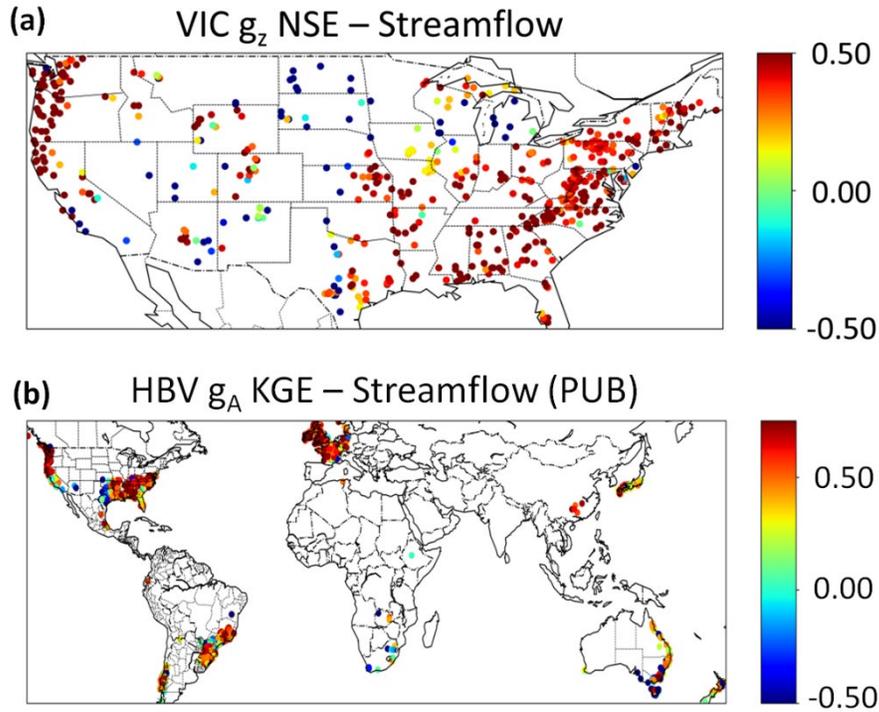

*Figure S4. Locations of basin outlets for (a) the Catchment Attributes and Meteorology for Large-Sample Studies (CAMELS). The color represents Nash Sutcliffe Model Efficiency coefficient (NSE) using parameters estimated by $g_z$ for the Variable Infiltration Capacity (VIC) hydrologic model; (b) The global headwater catchment database from Beck et al., 2020. The colors represent metric Kling-Gupta model efficiency coefficient (KGE) using parameters estimated by $g_A$ for the HBV hydrologic model.*

*Table S1.* Mean and standard deviation (std) of CONUS-scale evaluation metrics of the ensemble (with different random seeds) for default NLDAS-2 parameters, SCEUA-derived parameters, and dPL-derived parameters at $1/8^2$ sampling density. NLDAS-2 does not have an ensemble so the std is absent.

| Metric | Parameter source | SMAP (mean ± std) | SMAP spatial extrapolation (mean ± std) |
|---|---|---|---|
| CONUS-median Bias | NLDAS-2 s8 | 0.045 | |
| | SCE-UA s8 | -0.0012 ± 0.0001 | -0.0010 ± 0.0014 |
| | *dPL* s8 | 0.0016 ± 0.0026 | -0.0001 ± 0.0026 |
| | *dPL* s8A | -0.0011 ± 0.0019 | -0.0017 ± 0.0027 |
| CONUS-median Corr | NLDAS-2 s8 | 0.5292 | - |
| | SCE-UA s8 | 0.5891 ± 0.0025 | 0.5905 ± 0.0027 |
| | *dPL* s8 | 0.6000 ± 0.0318 | 0.6170 ± 0.023 |
| | *dPL* s8A | 0.5907 ± 0.0221 | 0.6020 ± 0.0158 |